\UseRawInputEncoding
\documentclass[twoside,11pt]{article}

\usepackage[cmex10]{amsmath}
\usepackage{amssymb}
\usepackage{algorithmic}
\usepackage[linesnumbered,ruled]{algorithm2e}
\usepackage{enumerate}
\usepackage{array}
\usepackage{color}
\usepackage{multirow}
\usepackage{adjustbox}
\usepackage[table,xcdraw]{xcolor}
\usepackage{textcomp}
\usepackage{booktabs,makecell}
\usepackage[justification=centering]{caption}
\usepackage{enumitem}
\usepackage{slashbox}
\usepackage{lipsum}
\usepackage{xcolor}
\usepackage{xparse}
\usepackage{jmlr2e}

\firstpageno{1}

\usepackage{lastpage}
\jmlrheading{22}{2021}{1-\pageref{LastPage}}{11/19; Revised 8/20}{1/21}{19-924}{ Imtiaz Ahmed, Xia Ben Hu, Mithun P. Acharya, Yu Ding} \ShortHeadings{Neighborhood Structure Assisted NMF for Anomaly Detection}{Ahmed, Hu, Acharya and Ding}

\begin{document}

\title{Neighborhood Structure Assisted Non-negative Matrix Factorization and Its Application in Unsupervised Point-wise Anomaly Detection}

\author{\name Imtiaz Ahmed \email imtiazavi@tamu.edu \\
       \addr Department of Industrial and Systems Engineering\\
       Texas A \& M University\\
       College Station, TX, USA
       \AND
       \name Xia Ben Hu \email hu@cse.tamu.edu\\
       \addr Department of Computer Science and Engineering\\
       Texas A \& M University\\
       College Station, TX, USA
       \AND
       \name Mithun P. Acharya \email mithun@totvs.com  \\
       \addr ABB Corporate Research\\
       Raleigh, NC, USA \\
       (Mithun Acharya is currently at TOTVS Labs. This work was done when he was at ABB Corporate Research)
              \AND
       \name Yu Ding \email yuding@tamu.edu \\
       \addr Department of Industrial and Systems Engineering\\
       Texas A \& M University\\
       College Station, TX, USA}

\editor{Qiaozhu Mei}

\maketitle

\begin{abstract}

Dimensionality reduction is considered as an important step for ensuring competitive performance in unsupervised learning such as anomaly detection. Non-negative matrix factorization (NMF) is a widely used method to accomplish this goal. But NMF do not have the provision to include the neighborhood structure information and, as a result, may fail to provide satisfactory performance in presence of nonlinear manifold structure. To address this shortcoming, we propose to consider the neighborhood structural similarity information within the NMF framework and do so by modeling the data through a minimum spanning tree. We label the resulting method as the neighborhood structure-assisted NMF. We further develop both offline and online algorithms for implementing the proposed method. Empirical comparisons using twenty benchmark data sets as well as an industrial data set extracted from a hydropower plant demonstrate the superiority of the neighborhood structure-assisted NMF. Looking closer into the formulation and properties of the proposed NMF method and comparing it with several NMF variants reveal that inclusion of the MST-based neighborhood structure plays a key role in attaining the enhanced performance in anomaly detection.

\end{abstract}



\section{Introduction}

Matrix factorization (MF) is one popular framework for finding the low dimensional embedding in a high dimensional data set. MF based approaches have been employed successfully to represent and group high dimensional data for better learning capability. Traditional matrix factorization produces low rank matrices consisting of negative values, and positive and negative weights, which tend to cancel each other in reconstructing the original matrix, and hence provide no intuitive meaning. The non-negative matrix factorization~\citep[NMF]{lee:1999} method, which imposes the non-negativity constraint in matrix factorization and only allows additive linear combinations of components, comes out as a better candidate for finding the low dimensional representation of high dimensional data. NMF has the capability of generating both clustering assignments and meaningful attribute distribution in two separate matrices. Immediately after its introduction, NMF not only becomes a powerful tool for clustering~\citep{xu:2003}, but it also shows enough potential in anomaly detection~\citep{allan:2008,tong:2011,liu:2017}.

In the presence of complicated manifolds, however, researchers notice that NMF starts to lose its efficacy~\citep{kuang:2012,cai:2011} as it only tries to approximate the data without trying to mimic the similarity among observations in the latent space. In other words, the shortcoming of the original NMF is attributed to that it has no provision to include the neighborhood structure information during the calculation of the factored matrices and thus cannot approximate the manifold embedded in the data.

Our research finds it beneficial to include the structural similarity information of data in the objective function of an NMF-based method, along with the original attribute information. Our specific approach is as follows. First, we convert the original data matrix into a graph object where each node represents an observation and each edge represents the virtual connection between a pair of data points. Then we apply a minimum spanning tree~\citep[MST]{prim:1957} on the graph to build a similarity matrix which is sparse, thus rendering computational efficiency. This MST-based similarity matrix also has the advantage to approximate the manifold structure of a local neighborhood~\citep{costa:2003}, better than simple Euclidean distances could.

We refer to the resulting method in this paper as the neighborhood structure-assisted NMF (NS-NMF). We demonstrate the benefit of the neighborhood structure-assisted NMF in the mission of anomaly detection. Our study shows that the proposed NS-NMF considers the local invariance property of data points while clustering data in a low-dimensional space, and doing so makes NS-NMF a powerful method for anomaly detection. Local invariance property ensures that the data points maintain the same neighborhood structure in the latent space, leading to better detection outcomes.

We want to note that two recent versions of the NMF method are closely related to what we propose in this paper. One of them is the graph regularized NMF~\citep[GNMF]{cai:2011}, which regularizes the original NMF formulation using a Laplacian matrix. Different from the proposed NS-NMF, GNMF constructs the similarity matrix based on simple Euclidean distances. Our numerical testing shows furthermore that GNMF, when employed for anomaly detection, is rather sensitive to two of its tuning parameters---the number of nearest neighbors and the regularization parameter. By contrast, NS-NMF does not need the neighborhood parameter and its detection outcome appears to be much less sensitive to its regularization parameter. Due to the popularity of GNMF, there exists multiple variants of the GNMF model~\citep{huang:2018, huang:2020}. However, they are all similar in terms of the strategy of capturing the neighborhood structure, which is using Euclidean distances.  Another NMF method that we want to specifically mention is the symmetric NMF~\citep[SNMF]{kuang:2012}, which uses only the similarity information while excluding the attribute information to generate the low rank matrices. In absence of the attribute information, SNMF depends on a dense pairwise similarity measure which leads to computational disadvantage. By abandoning the original attribute information in its formulation, SNMF makes its detection outcomes less interpretable than NS-NMF or GNMF. We conduct a comparison study in Section~\ref{section5.1} among all the NMF-based detection methods, i.e., the vanilla NMF, GNMF, SNMF, and NS-NMF, over 20 benchmark data sets. The comparison study shows a clear and evident advantage of NS-NMF.

To root our development in the background of anomaly detection, we would like to note that anomaly detection is related to \textit{clustering}, as researchers argue that detecting anomalies is to separate the data points into two classes---normal or regular versus abnormal, irregular, or anomalous~\citep{ester:1996, ertoz:2004, yu:2002, otey:2003, he:2003, amer:2012}.  As the NMF-based methods (or rather, all MF-based) try to cluster the data in the low-dimensional feature space, this branch of method falls naturally under the framework of subspace-based methods for anomaly detection~\citep{Zhang:2004, Kriegel:2009, zimek:2012, Muller:2008, keller:2012, van:2016}. We want to further note that NMF-based methods bear some conceptual similarity with the deep neural network (DNN)-based anomaly detection approaches~\citep{zhou:2017,zenati:2018,zong:2018,zhang:2019a}, as the DNN-based approaches also utilize the low-dimensional representation of data points for detecting anomalies. Unlike the linear low-dimensional transformation used by NMF-based methods, DNN-based approaches reach to the latent space through multiple steps of nonlinear transformation which could be more helpful in dealing with complex data structure like images. We provide some additional results in the Appendix, comparing our proposed approach with two DNN-based approaches.

The major contributions of our research reported here can be summarized as follows. First, we use an MST-based similarity matrix to capture the neighborhood structure and incorporate it into an NMF framework, so that the local structure is preserved during the low-dimensional transformation. Second, we provide an in-depth understanding of how the MST-based similarity measure makes a difference for anomaly detection through an analysis of the formulations and properties of the NMF variants, including our own NS-NMF. Third, we propose an online version of the NS-NMF approach which can be applied to streaming data and help the practitioners in real-time anomaly detection.  The take-home message is that after dimensionality reduction via NMF, the performance of anomaly detection can be much enhanced when a MST-based similarity measure is used. 

Our previous effort~\citep{ahmed:2019a} also showed that making use of the neighborhood structure through an MST model helps the objective of anomaly detection in general, even without dimensionality reduction, as this previous effort,~\citet{ahmed:2019a}, works as a stand alone anomaly detection approach and is applied to the original data. The method in~\citet{ahmed:2019a} falls under the framework of neighborhood methods, whereas the NMF-based methods are under the framework of subspace methods. In the literature of anomaly detection, the two frameworks of methods are considered in parallel. There is no general consensus to assert which framework is definitely better, and if a strategy works better under one framework, there is no guarantee that it would work as well under the other framework.  We will nonetheless present a comparison between the NMF-based anomaly detection and a non-NMF method in Section~\ref{section5.1b}.

The rest of the paper unfolds as follows. Section~\ref{sec:formulation} describes the detailed formulation of the NS-NMF. Section~\ref{sec:compare} discusses the similarities and differences among the proposed NS-NMF, GNMF and SNMF. Section~\ref{sec:algorithm} presents the proposed NS-NMF algorithm in a structured way for both offline and online versions. Section~\ref{sec:performance} empirically compares the proposed NS-NMF method with other NMF variants and a non-NMF approach using 20 benchmark data sets.  We also apply these methods to a hydropower data set. Finally, we summarize the paper in Section~\ref{sec:conclusion}.

\section{Incorporating Neighborhood Structure Information into NMF}\label{sec:formulation}
Anomaly detection is by and large an unsupervised learning problem as the class labels of data records are unknown in the training set. One has to depend on the structure of the data to flag a potential anomaly.  Anomalies could be \emph{global} and lie far away from most of the data points, thus making it easy to identify them, or it could be \emph{local}, homogeneously co-exist around the regular data points, and only be found if compared with proper neighboring sets/clusters. Nonetheless, to differentiate the anomalies from the normal data points, one may need to solve two problems. The first is to find the appropriate local contexts/communities as the latent feature groups and the second is to extract the characteristics of these communities in terms of the features. We can reconstruct the original data points using the combination of the two low-dimensional representations. The reconstruction can then be used to judge a data point for flagging and short listing potential anomalies. NMF apparently provides an effective solution to both of these problems. However, as we have pointed out earlier, the traditional NMF framework does not take into consideration the neighborhood structure of data points while conducting clustering in the latent space. It will thereby produce unsatisfactory results in the presence of complicated manifolds. We try to bridge this gap by proposing a graph-based approach, to be used together with NMF, to account for the existence of neighborhood structures. The new formulation considers the local invariance property while obtaining the low-dimensional representation, and thus works better than the traditional approaches.

\subsection{Basic NMF Framework}

In NMF~\citep{lee:1999}, a data matrix $\mathbf A \in \mathbb{R}_{+}^{n\times p}$, of which columns and rows represent the attributes and observations respectively, is factorized into two low rank matrices, namely, $\mathbf W \in \mathbb{R}_{+}^{n\times K}$ and $\mathbf H \in \mathbb{R}_{+}^{K\times p}$, such that the inner product of these factorized matrices approximate the original data matrix. Here, $K$ represents the number of latent feature groups and it is required to be equal or less than the smaller of $n$ and $p$. NMF tries to project the data with high dimensional features into a low dimensional latent space so that the original observations can be seen as a weighted linear combination of the newly formed basis vectors corresponding to each latent feature group. The rows of $\mathbf H$, each of which is a $1\times p$ vector, represent the basis vectors, whereas the weights come from $\mathbf W$. For example, any row $i$  from the original matrix can be reconstructed as
\begin{equation}
\label{eq:apx}
\mathbf{a}_{i} = \sum_{k=1}^K W_{ik} \mathbf{h}_{k},
\end{equation}
where $\mathbf a_i$ represents the $i$-th row of $\mathbf A$, $\mathbf h_k$ represents the $k$-th row of $\mathbf H$, and $W_{ik}$ is the $(i,k)$-th element of $\mathbf W$.

To solve for the factored matrices, one needs to minimize the Frobenius norm of the difference between the original data matrix and the inner product of the factored matrices, as shown in Equation \ref{eq:frob}, i.e.,
\begin{equation}
\label{eq:frob}
\min_{\mathbf{W\geqslant0, \, H\geqslant0}}\left\Vert \mathbf{A - WH} \right\Vert^2_F.
\end{equation}
where $\Vert \cdot\Vert_F$ denotes the Frobenius norm and the constraints, $\mathbf{W\geqslant0 \text{ and } H\geqslant0}$, mean that both $\mathbf W$ and $\mathbf H$ are non-negative matrices.

\subsection{Capturing Neighborhood Structure Using MST}

NMF finds its way in solving clustering problems when~\cite{ding:2005} show that NMF can be made equivalent to clustering approaches by reframing the problem slightly. While doing clustering, however, NMF does not consider neighborhood structure information. But the neighborhood structure information should have been considered for the benefit of clustering or detection, as structurally similar observations in the original space ought to maintain the similarity in the latent space. To tackle this problem, we propose to extract the neighborhood structure information from the original data matrix via the modeling of a minimum spanning tree and then incorporate the structure information during the calculation of the NMF factored matrices.

To discover the intrinsic structure in data, a popular undertaking is to form a graph object using the original data matrix $\mathbf A$. Each observation is represented by a node, which is connected with other nodes through a weighted edge with the weight being the pairwise Euclidean distance between them.  A simple graph like this has its disadvantage---the similarity matrix thus generated would be too dense to be incorporated in the NMF setting for large data sets. In our treatment, we instead use the MST for constructing the similarity matrix among data points, and doing so leads to a sparse similarity matrix.

\begin{figure}[t]
	\centering
	\centerline{\includegraphics[width=5in, height=3in]{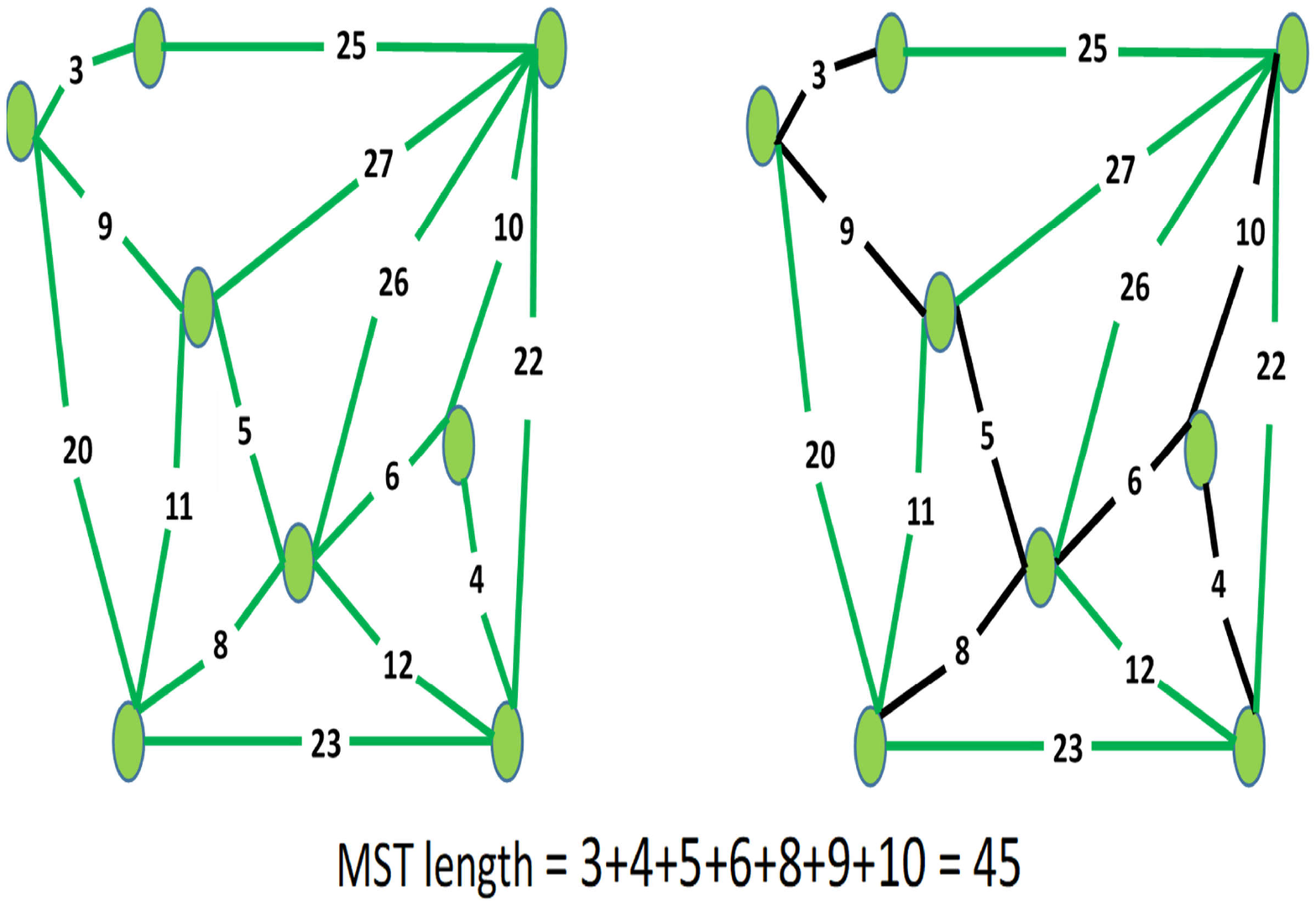}}
           \caption{Formation of an MST: the left panel is the initial graph, and the black colored edges form the minimum spanning tree in the right panel.}
	\label{fig:formationMST}
\end{figure}

To understand the concept of MST, consider a connected undirected graph $G = (V, E)$, where $V$ is the collection of nodes and $E$ represents the collection of edges connecting these vertices as pairs. For an edge $e \in E$, as mentioned above, a weight is associated with it, which is the pairwise Euclidean distance between the chosen pair of nodes. A minimum spanning tree is a subset of the edges in $E$ that connects all the vertices together, without any cycles and with the minimum possible total edge weight. Consider the example in Fig.\ref{fig:formationMST}, left panel, where there are 8 nodes and 15 edges connecting the nodes. Each of the edges has a unique edge length associated with it which is represented by a numeric value. If we want to connect all the nodes using the given edges without forming a cycle, there could be many such combinations with only one having the minimum total edge length, which is shown in the right panel. The edges in black color represent the selected 7 edges from the 15 total edges. The resulting graph in the right panel, consisting of the black edges only, is the MST for the initial connected graph. Apparently, MST compresses the original graph and preserves certain degree of information that we consider important for anomaly detection purpose. 

Once we apply MST on the graph object resulting from the original data matrix $\mathbf A$ having $n$ observations, what we get is a square matrix, $\mathbf M \in \mathbb{R}_{+}^{n\times n}$, showing the pairwise connectedness and distance. A strictly positive value in $\mathbf M$ represents the distance between two connected nodes in the resulting MST, whereas a zero implies that the nodes are not adjacent. Unlike for a complete graph, $\mathbf M$ of MST is supposed to be a sparse matrix, rather than a dense matrix.  We further convert it into a pairwise similarity matrix, $\mathbf S$, by inverting only the positive entries of $\mathbf M$, as following:
\[
    S_{ij}=
\begin{cases}
    \frac{1}{M_{ij}},& \text{if } M_{ij}>0,\\
    0,              & \text{otherwise},
\end{cases}
\]
where $S_{ij}$ and $M_{ij}$ are, respectively, the $(i,j)$-th elements of $\mathbf S$ and $\mathbf M$. Note that $S_{ii} = 0$ because $M_{ii}=0$. This $\mathbf S$ matrix is called a similarity matrix because a high value in $\mathbf S$ is the result of two nodes close to each other in the resulting MST, implying their similarity and likely the same cluster membership.  On the other hand, a zero value means that two nodes are not directly connected and less likely similar to each other.  Whether or not they may still belong to the same cluster depends largely on the two nodes' association with the common neighbors.  Understandably, if two points have very low similarity and do not have connections through any common neighbors, they most probably belong to two separate clusters. We intend for $\mathbf S$ to guide the basic NMF process to group similar observations into the same cluster, obtain the proper cluster centroids in the form of basis vectors, and subsequently use both of them in the action of anomaly detections.

\subsection{The NS-NMF Formulation}

By taking into account both the original attribute matrix $\mathbf A$ and the MST based neighborhood similarity matrix $\mathbf S$, we adopt the following generalized NMF formulation~\citep{liu:2017}:
\begin{equation}
\label{eq:main}
\min_{\mathbf{W\geqslant0, \, H\geqslant0}}\left\Vert \mathbf{S - WW}^T \right\Vert^2_F + \alpha\left\Vert \mathbf{A - WH} \right\Vert^2_F + \gamma(\left\Vert \mathbf{W} \right\Vert^2_F + \left\Vert \mathbf{H} \right\Vert^2_F) ,
\end{equation}
where $\left\Vert \mathbf{W} \right\Vert^2_F$ and $\left\Vert \mathbf{H} \right\Vert^2_F$ are the regularization terms added to the objective function to avoid overfitting, $\gamma$ is the regularization parameter controlling the extent of overfitting. The first term is apparently the newly added structure similarity term, whereas the second term is the original NMF cost function, and the parameter $\alpha$ is used to trade off between these two cost functions and typically takes values between 0 and 1.  The reason that $\alpha$ is restricted to be no greater than one is because a greater $\alpha$ pushes the resulting method towards the plain version of NMF. Analogously, in Equation \ref{eq:GNMF} below, the counterpart parameter in GNMF is $\lambda$, which is generally assigned a value greater than one, although it can technically take any nonnegative values.

As we will show in Section~\ref{sec:compare}, the expression in Equation~\ref{eq:main} is a generalized and unified formulation for many variants of NMF methods.  For instance, by setting $\alpha$ really large and $\gamma=0$, the generalized formulation reduces to the vanilla NMF. Section~\ref{sec:compare} provides details regarding how GNMF and SNMF can also be derived from this formulation.  While this generalized formulation does provide a unified framework, allowing us to understand better the relationship among several NMF methods, we want to stress that the uniqueness of NS-NMF lies in the specific formation of its similarity matrix, $\mathbf S$, as its $\mathbf S$ is derived from an MST representing the underlying data structure.

This unique way of forming the similarity matrix in NS-NMF makes a profound difference.  Recall the vanilla NMF only tries to find the basis vectors that best approximate the data. While the vanilla NMF can achieve successes in the presence of simple Euclidean structure, it does not appear effective in the presence of complicated intrinsic data structure. In the latter cases the vanilla NMF may achieve the data approximation objective but the clustering weights may not be appropriate.  But having proper clustering weights is important for determining appropriate community association and anomaly identification. As we will see in Equation \ref{eq:anomaly} later, $W_{ik}$ plays a key role in computing the anomaly score along with the basis vector $\mathbf h_k$.

In a curved space (like the surface of the earth), or more generally speaking, in a structured space, geodesic distances measure the minimum possible distance between two points. The MST-based neighborhood structure is able to approximate the geodesic distance between data instances via a multi-hop edge connection on the tree graph, which are considered a much better representation of complicated data structures than simple Euclidean distances~\citep{Meichen:1995,Tu:2016,costa:2003}. By incorporating the MST-based similarity measure in NMF, the resulting method produces better clustering assignments with respective cluster centroids extracted from $\mathbf H$; doing so, we believe, enhances substantially the ability of anomaly detection.

\subsection{Anomaly Detection}

Now, let us take a look at how the proposed NS-NMF can help us in distinguishing anomalies from the normal observations. The low rank factored matrices generated from NS-NMF provide us with the information pertinent to anomaly detection. Each entry of $\mathbf W$ measures the extent of an observation's association with all $K$ latent groups/clusters, whereas a row of $\mathbf H$ represents the average characteristics of one of the latent groups/clusters. Together, they can reconstruct the original observations. Reconstruction error for each observation measures the quality of the reconstruction, which can be in turn used as an indicator of the degree of anomalousness associated with an observation.

If we represent the reconstruction as $\mathbf A^\prime$, then the reconstruction error is quantified as the loss between $\mathbf A$ and $\mathbf A^\prime$, such as
\begin{equation}
\label{eq:loss1}
L(\mathbf A,{\mathbf A}')= \left \|\mathbf A-{\mathbf A}'  \right \|_{F}=\left \|\mathbf A-\mathbf{WH} \right \|_{F}.
\end{equation}
NMF variants, or even broadly, any dimensionality reduction approaches, are to encode the observations in a lower-dimensional latent space while preserving the main characteristics of the data. The quality of such preservation action is measured by the above reconstruction error.

An anomaly detection procedure based on the reconstruction error assumes that anomalies do not belong to the regular data distribution, and therefore, it is harder to reconstruct them from their compressed low dimensional representation compared to the normal data points. As a result, anomalies will most likely exhibit a higher reconstruction error. The higher error associated with an observation, the higher chance for it being anomalous.

Specifically, the reconstruction error-based anomaly score is thus computed after applying the NS-NMF procedure. In NS-NMF, we try to encode the learning of an effective low-dimensional representation into the two low-rank matrices, $\mathbf W$ and $\mathbf H$.  The second part of the objective function in Equation \ref{eq:main} makes sure that the difference between the observations in $\mathbf A$ and its reconstruction $\mathbf {WH}$ remains as small as possible, whereas the first part of Equation \ref{eq:main} makes sure that observations which are close in the original space maintains the relative closeness in the latent space. At the end of the low-dimensional representation learning, we compute the reconstruction error for individual observations as
\begin{equation}
\label{eq:anomaly}
O_{i} = \left \|\mathbf {a}_{i}-{\mathbf {a}_{i}}'  \right \|_{2} = \left \|\mathbf {a}_{i}-{\sum_{k=1}^K W_{ik} \mathbf{h}_{k}}  \right \|_{2},
\end{equation}
and then repeat this process for all observations. Finally, rank the scores, $\{O_i, i=1, \ldots, n\}$, in descending order.  Those observations associated with high anomaly scores are flagged as anomalous.

In the practice of anomaly detection, it is common that once applied to the data, a method ranks the top $N$ instances as potential anomalies and treats the rest of data instances as normal instances. One main reason for such a decision process is that unsupervised anomaly detection methods tend to have a lower detection capability and higher false alarm rate. As a result, unsupervised detection methods are typically used as a screening tool, flagging potential anomalies to be further analyzed by either a human operator or some more expensive procedure. A cut-off is therefore used to ensure the subsequent step---more expensive or time consuming---practical and feasible. In this paper, we follow this practice to declare the observations with top $N$ scores as anomalies where the cut-off threshold, $N$, is prescribed based on the cost/feasibility considerations.


\section{NS-NMF Relative to Other NMFs}\label{sec:compare}

In this section, we want to highlight the similarities and contrast of the proposed NS-NMF with some of the related approaches, principally GNMF and SNMF so that the readers get a better understanding of how the proposed method made the difference.

First, we present the formulations of the three methods in Equations \ref{eq:GNMF}--\ref{eq:NSNMF}. To capture the essence of the NS-NMF method, we rewrite the formulation in Equation \ref{eq:main} by ignoring the third component of the objective function, because the third term is a regularization term to avoid overfitting, and as such, having it or not does not change the essence concerning which piece of information is used in the matrix factorization. The new formulation of NS-NMF in Equation \ref{eq:NSNMF} now has two terms---this can be viewed as setting $\gamma=0$. We also change the position of $\alpha$ for easiness of comparison.
\begin{align}
\text{GNMF}&:\min_{\mathbf{W,H\geqslant0}}\lambda \cdot \text{tr}(\mathbf{W}^T\mathbf{LW})  + \left\Vert \mathbf{A - WH} \right\Vert^2_F. \label{eq:GNMF} \\
\text{SNMF}&:\min_{\mathbf{W\geqslant0}}\left\Vert \mathbf{S - WW}^T \right\Vert^2_F. \label{eq:SNMF}\\
\text{NS-NMF}&:\min_{\mathbf{W,H\geqslant0}}\frac{1}{\alpha}\cdot\left\Vert \mathbf{S - WW}^T \right\Vert^2_F + \left\Vert \mathbf{A - WH} \right\Vert^2_F. \label{eq:NSNMF}
\end{align}

Conceptually and formulation wise, GNMF is the closest to the proposed NS-NMF. We notice that both NS-NMF and GNMF formulations have the same second component. This second component comes from the vanilla NMF formulation and is used to obtain two factor matrices from the original data matrix, one of which is the attribute matrix.

Admittedly, the authors of GNMF are the first to shed light on the necessity of considering neighborhood similarity information in the NMF process. According to \citet{cai:2011}, in order to incorporate the neighborhood similarity information, the low rank approximation should be obtained as in Equation \ref{eq:GNMF}, rather than in Equation \ref{eq:frob}. The specific mechanism of incorporating the neighborhood similarity information in Equation \ref{eq:GNMF} is through the use of a graph Laplacian matrix, denoted by $\mathbf L$.  The graph Laplacian matrix $\mathbf L$ can be obtained by $\mathbf L =\mathbf D - \tilde{\mathbf S}$, where $\mathbf D$ is a diagonal matrix also known as the degree matrix~\citep{cai:2011,ding:2005} and $\tilde{\mathbf{S}}$ is the adjacency matrix. The adjacency matrix is calculated using the Euclidean structured neighborhood information after converting the original data into a graph object. The model has two main parameters, namely, $q$, the number of the nearest neighbors to be specified in order to form a similarity matrix, and, $\lambda$, the regularization parameter. The value of $\lambda$ can take any non-negative value. When it takes the value of zero, the formulation ignores the neighborhood similarity completely and GNMF reduces to the vanilla NMF.

SNMF, on the other hand, promotes the idea that one should probably consider the neighborhood similarity information only while obtaining the factored matrices. On surface, SNMF can be seen as a special case of the generalized formulation in Equation~\ref{eq:main} that sets $\alpha=0$ and $\gamma=0$. The factorization of the similarity matrix $\mathbf S$ generates a clustering assignment matrix $\mathbf W$ that is also constrained to be non-negative. The authors of SNMF believe that doing so captures the inherent cluster structure from the graph representation of the original data matrix. \citet{kuang:2012} argue, and present case studies in support of, that the traditional NMF is not ideal for handling data sets with nonlinear structures. Compared to the vanilla NMF, SNMF can deal with complicated patterns and generate more accurate clustering assignments. \citet{kuang:2012} also point out that the SNMF formulation is in fact equivalent to some graph clustering methods including spectral clustering~\citep{ding:2005}. \citet{kuang:2012} present an example to show that adding the non-negativity constraints help the clustering objective and also that SNMF performs more robustly compared to other clustering methods, as SNMF does not depend on the eigenvalue structure which tends to provide inaccurate results if certain conditions are violated.

SNMF formulation takes basically the first component of the NS-NMF formulation, while ignoring the vanilla NMF portion. In this way, SNMF focuses on the accuracy of clustering assignments.  But for the mission of anomaly detection, we need something more. What we need is to have a basis connecting the latent features with the original ones, so that we can reconstruct observations from the low-dimensional representations and pinpoint anomalies by looking at their deviation from the original form. The second portion of the NS-NMF formulation helps us obtain a basis matrix which summarizes the attribute distribution of the latent feature groups, thereby providing the means for detecting the anomalies from the reconstruction. For this reason, NS-NMF, which keeps the second term, is more meaningful for anomaly detection than SNMF. Even the first term, although appears the same in both SNMF and NS-NMF formulations, is not really the same.  The difference of SNMF and NS-NMF in their first terms, less apparent but arguably more critical, is that they use the different similarity matrix $\mathbf S$ (but the mathematical notation looks the same).  SNMF uses the traditional Euclidean distance-based similarity metric considering the full graph, while that in NS-NMF comes from an MST. The similarity matrix in SNMF is too dense, making SNMF to suffer in case of approximating complex structures. Unsurprisingly, SNMF is also computationally more expensive than NS-NMF.

Coming back to GNMF, which has the same second term as NS-NMF and poised to be more suitable for anomaly detection.  The first term in both NS-NMF and GNMF formulations has a strong connection.  It can be shown that when using the same similarity matrix $\mathbf S$, GNMF and NS-NMF can be made (nearly) equivalent.

To facilitate the understanding of this connection, let us consider adding an orthogonality constraint on $\mathbf{W}$.  This is not exactly required in the original formulations but \citet{ding:2005} shows that minimizing $\left\Vert \mathbf{S - WW}^T \right\Vert^2_F$ retains the orthogonality of $\mathbf{W}$ approximately. Furthermore, suppose that a symmetric normalized Laplacian matrix is used, i.e., the original $\mathbf L$ is pre- and post-multiplied by $\mathbf D^{-\frac{1}{2}}$. Then we have
\[\mathbf{L} = \mathbf{I} - \mathbf{D}^{-\frac{1}{2}}\tilde{\mathbf S}\mathbf{D}^{-\frac{1}{2}},\]
where without ambiguity, we still use $\mathbf L$ to denote the symmetric normalized Laplacian matrix. Denote by $\mathbf{S} =\mathbf{D}^{-\frac{1}{2}}\tilde{\mathbf S}\mathbf{D}^{-\frac{1}{2}}$ the newly generated normalized similarity/adjacency matrix. As such, the first term in GNMF can be made equivalent to that of NS-NMF by considering the following minimization formulation.
\begin{equation}\label{eq:equivalence}
\begin{split}
\min_{\mathbf{W\geqslant0}} \text{tr}(\mathbf{W}^T\mathbf{LW}) &=  \min_{\mathbf{W\geqslant0}}\text{tr}(\mathbf{W}^T(\mathbf{I} - \mathbf{D}^{-\frac{1}{2}}\tilde{\mathbf S}\mathbf{D}^{-\frac{1}{2}})\mathbf{W})\\
& =  \min_{\mathbf{W\geqslant0}}\text{tr}(\mathbf{W}^T(\mathbf{I} - \mathbf{S})\mathbf{W})\\
& = \min_{\mathbf{W\geqslant0}} \text{tr}(\mathbf{W}^T\mathbf{W}) -\text{tr}(\mathbf{W}^T\mathbf{SW}) \\
& = \min_{\mathbf{W\geqslant0}; \mathbf{W}^T\mathbf{W}=\mathbf{I}} \text{tr}(\mathbf I) -\text{tr}(\mathbf{W}^T\mathbf{SW}).
\end{split}
\end{equation}
At the last expression, we use the orthogonality constraint on $\mathbf{W}$, as mentioned above. Minimizing Equation \ref{eq:equivalence} with respective to a non-negative $\mathbf W$ does not change its minimization outcome if we add a term that does not depend on $\mathbf W$ or if we multiply the $\mathbf W$-depending term by a constant.  Let us then add one term, $\text{tr}(\mathbf{S}^T\mathbf{S})$, to the last expression of Equation \ref{eq:equivalence} and multiply $\text{tr}(\mathbf{W}^T\mathbf{SW})$ by two.  As such, we end up with an equivalent minimization problem as follows:
\begin{equation}\label{eq:equivalence2}
\begin{split}
\min_{\mathbf{W\geqslant0}} \text{tr}(\mathbf{W}^T\mathbf{LW}) \quad \text{is equivalent to} \quad &\min_{\mathbf{W\geqslant0};\mathbf{W}^T\mathbf{W}=\mathbf{I}}\text{tr}(\mathbf{I}) -2\text{tr}(\mathbf{W}^T\mathbf{SW}) + \text{tr}(\mathbf{S}^T\mathbf{S})\\
& = \min_{\mathbf{W\geqslant0};\mathbf{W}^T\mathbf{W}=\mathbf{I}}\text{tr}[(\mathbf{S - WW}^T)^T(\mathbf{S - WW}^T)] \\
& = \min_{\mathbf{W\geqslant0};\mathbf{W}^T\mathbf{W}=\mathbf{I}}\left\Vert \mathbf{S - WW}^T \right\Vert^2_F.
\end{split}
\end{equation}

The above derivation makes it apparent that if one uses the same similarity matrix $\mathbf S$ in both GNMF and NS-NMF formulations, then GNMF is practically the same as NS-NMF.  Of course, which $\mathbf S$ to use creates the difference between GNMF and NS-NMF.  GNMF uses only a fixed small subset of the neighborhood/adjacency information to obtain an Euclidean distance-based similarity matrix and then convert it to a graph Laplacian form. A prescribed neighborhood size, $q$, is one of the parameters used in GNMF. NS-NMF's similarity matrix, on the other hand, is based on an MST and differs from that of GNMF. NS-NMF does not need the neighborhood size parameter, due to its use of MST. Both methods use a regularization parameter, which is $\alpha \, (0-1)$ in NS-NMF and $\lambda \, (\geq 0)$ in GNMF; these regularization parameters are in fact equivalent. Our numerical analysis shows that GNMF is sensitive to both of its parameters, $q$ and $\lambda$, while the NS-NMF is reasonably less sensitive to its parameter $\alpha$. We believe that this is a benefit of using the MST-based similarity matrix.

\section{Algorithmic Implementation of NS-NMF}\label{sec:algorithm}

In this section, we discuss the implementation of the NS-NMF proposed in Section~\ref{sec:formulation}.  We provide two algorithmic procedures, one is an accelerated offline implementation and the other is an online implementation, i.e., processing observations one at a time for real-time anomaly detection.

\subsection{Accelerated Offline Implementation}

A number of algorithms have already been proposed to solve the original NMF problem and its variants. However, most of them are computationally expensive and not ideal for handling big data scenarios. In this paper, we choose to utilize a distributed version of the stochastic gradient descent (SGD) algorithm ~\citep{liu:2017,gemulla:2011} which enables us to achieve an accelerated and parallel optimization scheme. In the traditional SGD, one updates the parameters at each round by going through a single training point at a time, whereas in the distributed version, we update the parameters by processing multiple independent blocks of training data in parallel and thereby take the computational advantage. Here, independence means that the parameter update of one block will not affect the parameter update of any other blocks. This property is also known as the interchangeable property~\citep{liu:2017,gemulla:2011}. To design a distributed SGD, we define a loss function as in Equation \ref{eq:loss}, which is just the blockwise summation of the objective function defined in Equation \ref{eq:main}, such as

\begin{equation}
\begin{aligned}
\label{eq:loss}
L_{\mathbf{S,A}} (\mathbf{W,H}) &=  \sum_{\{i,j,k\}} \Big(\left \| \mathbf{S}^{ij}-\mathbf{W}^{i}\mathbf{W}^{{j}^{T}}\right \|^{2}_F +\frac{\gamma }{2B} \left \| \mathbf{W}^{i}\right \|^{2}_F +\frac{\gamma }{2B} \left \|\mathbf{ W}^{j}\right \|^{2}_F\\
& + \frac{\alpha}{2}\left \| \mathbf{A}^{ik}-\mathbf{W}^{i}\mathbf{H}^{{k}^{T}}\right \|^{2}_F      + \frac{\alpha}{2}\left \| \mathbf{A}^{jk}-\mathbf{W}^{j}\mathbf{H}^{{k}^{T}}\right \|^{2}_F+\frac{\gamma }{B} \left \| \mathbf{H}^{k}\right \|^{2}_F\Big) \\
& = \sum_{\{i,j,k\}} L_{i,j,k} (\mathbf{W}^{i},\mathbf{W}^{j},\mathbf{H}^{k}),
\end{aligned}
\end{equation}
where $B$ represents the number of splits in each dimension and it controls the number of blocks created in $\mathbf S$ and $\mathbf A$. We essentially divide $\mathbf S$ and $\mathbf A$ into blocks and the position of each block is represented by the superscripts, $i,j,k$. Suppose that we have a $100\times100$ matrix and $B =5$. Then we will have 5 blocks with each containing 400 (i.e., a $20\times 20$ matrix) training points. To process and approximate a block $\mathbf{S}^{ij}$, we need to update parameter block $\mathbf{W}^{i}$ and $\mathbf{W}^{j}$. Likewise, to approximate a block of $\mathbf{A}^{ik}$, we need to update $\mathbf{W}^{i}$ and $\mathbf{H}^{k}$. As we have to process blocks from two separate matrices $\mathbf S$ and $\mathbf A$, with parameters to be updated connecting each other, we define the blocks to be processed from these two matrices as an instance set $\{\mathbf{S}^{ij}, \mathbf{A}^{ik}, \mathbf{A}^{jk}\}$.

To achieve distributed and parallel processing, we need to process in parallel the randomly generated instance sets at each round. It is possible to do so only if they are interchangeable and do not affect the resulting parameter updates $\{\mathbf{W}^{i}, \mathbf{W}^{j}, \mathbf{H}^{k}\}$ of one another.  According to \citet{gemulla:2011}, the interchangeability occurs only when the superscripts of the blocks do not coincide. For example, $\{\mathbf{S}^{15}, \mathbf{A}^{13}, \mathbf{A}^{53}\}$ and $\{\mathbf{S}^{23}, \mathbf{A}^{24}, \mathbf{A}^{34}\}$ are two interchangeable instances as they have entirely different superscripts. Now, as we have defined both the loss function and the instance sets, parameter update can be calculated according to Equation \ref{eq:update}, where $\mathbf{\theta}^{i,j,k}=\{\mathbf{W}^{i}, \mathbf{W}^{j}, \mathbf{H}^{k}\}$ and $\epsilon_{t}$ is the step size at current iteration.
\begin{equation}
\label{eq:update}
{\theta_{t+1}}^{i,j,k}  =  {\theta_{t}}^{i,j,k} - \epsilon_{t}\Delta L_{i,j,k} ({\theta_{t}}^{i,j,k}).
\end{equation}

The algorithm steps is summarized in Algorithm~\ref{algorithm1}, including the construction of MST-based similarity matrix, the NMF optimization procedure, and the detection of anomalies.

\begin{algorithm}[tb]
\caption{Offline implementation of NS-NMF algorithm for anomaly detection}
\label{algorithm1}
    \SetKwInOut{Input}{Input}
    \SetKwInOut{Output}{Output}
    \Input{$\mathbf{A}$, $\alpha$, $N$, $\gamma$, $B$, $K$}
    \Output{List of anomalous nodes $l_{final}$}
     Generate a set of vertices $V$, where each vertex represent a separate observation from the data set\;
     Construct edges by calculating Euclidean distance between each pair of vertices using their attribute values from the data set and store them in  $E$\;
     Construct a MST using $V$ and $E$ and generate the pairwise similarity matrix $\mathbf{S}$ from the resultant pairwise MST distance matrix\;
     Initialize $\mathbf{W}$ and $\mathbf{H}$ randomly\;
     Partition $\mathbf{S}$ and $\mathbf{A}$ and corresponding $\mathbf{W}$ and $\mathbf{H}$ into blocks\;
    \While{not converged} {
        Randomly generate a collection of instance sets from blocked $\mathbf{S}$ and $\mathbf{A}$, $U =\{\{i_{1},j_{1},k_{1}\},\{i_{2},j_{2},k_{2}\},\{i_{3},j_{3},k_{3}\},.....\}$ such that any two are interchangeable\;
        \For{$(i,j,k) \in U$ in parallel}{
        $\mathbf{W}^{'i} \gets \mathbf{W}^{i} - \epsilon_{t}\Delta_{\mathbf{W}^{i}} L_{i,j,k} $\;
         $\mathbf{W}^{'j} \gets \mathbf{W}^{j} - \epsilon_{t}\Delta_{\mathbf{W}^{j}} L_{i,j,k}$ (if  $i\neq j$)\;
         $\mathbf{H}^{'k} \gets \mathbf{H}^{k} - \epsilon_{t}\Delta_{\mathbf{H}^{k}} L_{i,j,k} $\;
        $\mathbf{W}^{i} \gets \mathbf{W}^{'i}, \mathbf{W}^{j} \gets \mathbf{W}^{'j},\mathbf{H}^{k} \gets \mathbf{H}^{'k}$\;
        Non negativity projection for $\mathbf{W}^{i}, \mathbf{W}^{j}$ and $\mathbf{H}^{k}$\;
        }
     }
     Calculate the anomaly scores of the observations according to Equation \ref{eq:anomaly} and store them in $O_{i}$ \;
Store the accumulated list of nodes with anomaly scores from all the clusters as $l_{total}=\{O_{1},O_{2},\ldots,O_{n}\}$\;
Return the nodes associated with top $N$ anomaly scores as final anomalies in $l_{final}$\;
\end{algorithm}

\subsection{Online Implementation}

The offline implementation discussed above has the advantage of parallel blockwise implementation but lacks the ability of instantaneous update and real-time anomaly score computing. It requires to see the data all at once and builds the MST using all the instances even before the execution of the algorithm. To evaluate an observation in real time, we need to find a way to update both weight matrix $\mathbf{W}$ and basis matrix $\mathbf{H}$ incrementally when new samples arrive in a streaming fashion. In addition, we need to make sure that such an update will not require the entire data matrix.

There has been quite a few efforts in developing the online version of the vanilla NMF~\citep{guan:2012,zhao:2016,zhu:2017,tu:2018,guo:2019}, although comparatively fewer attempts have been made for online GNMF~\citep{liu:2016}. It is so because adding geometric structures to guide the NMF process makes the online update more difficult as we need to calculate the geometric weights on the fly. In this section we layout the online implementation of NS-NMF.

Let us assume that the observations, $\mathbf A=[ \mathbf a_{1}, \mathbf a_{2},\ldots, \mathbf a_{d-1}, \mathbf a_{d}]$, are generated in a streaming fashion, where $\mathbf{a}_{d}$ represents the $d$th data sample just arrived and its attribute information. Upon its arrival, the $d$th component of the weight matrix and basis matrix need to be updated so that the instantaneous anomaly score can be calculated. For this $d$th data sample we can write our NS-NMF objective function as follows, which ignores the regularization component.
\begin{equation}
\begin{aligned}
\label{eq:stream}
\mathbf{F}_{d} &=\alpha\left\Vert \mathbf{A}_{d}- \mathbf W_{d} \mathbf H_{d} \right\Vert^2_F + \left\Vert \mathbf{S}_{d} - \mathbf W_{d}\mathbf W_{d}^T \right\Vert^2_F, \\
& \text{which is equivalent to, (recall Equation~\ref{eq:equivalence2})}\\
&\alpha\left\Vert \mathbf{A}_{d}- \mathbf W_{d}\mathbf H_{d} \right\Vert^2_F + \text{tr}(\mathbf{W}_{d}^T\mathbf{L}_{d}\mathbf W_{d})\\
&=\alpha\sum_{i=1}^{d}\sum_{j=1}^{p}((\mathbf{A}_{d})_{ij}-(\mathbf W_{d}\mathbf H_{d})_{ij})^{2} +\sum_{k=1}^{K}\sum_{i=1}^{d}\sum_{j=1}^{d}(\mathbf W_{d}^T)_{ki}(\mathbf L_{d})_{ij}(\mathbf W_{d})_{jk}\\
&=\Bigg[\alpha\sum_{i=1}^{d-1}\sum_{j=1}^{p}((\mathbf{A}_{d-1})_{ij}-(\mathbf W_{d-1}\mathbf H_{d})_{ij})^{2} +\sum_{k=1}^{K}\sum_{i=1}^{d-1}\sum_{j=1}^{d-1}(\mathbf W_{d-1}^T)_{ki}(\mathbf L_{d})_{ij}(\mathbf W_{d-1})_{jk}\Bigg]\\
&+ \Bigg[\alpha\sum_{j=1}^{p}((\mathbf a_{d})_{j}-(\mathbf w_{d}\mathbf H_{d})_{j})^{2} + \sum_{k=1}^{K}\sum_{i=1}^{d-1}(\mathbf W_{d}^T)_{ki}(\mathbf L_{d})_{id}(\mathbf w_{d})_{k} + \sum_{k=1}^{K}\sum_{j=1}^{d-1}(\mathbf w_{d}^T)_{k}(\mathbf L_{d})_{dj}(\mathbf W_{d})_{jk}\\
&+ \sum_{k=1}^{K}(\mathbf w_{d}^T)_{k}(\mathbf L_{d})_{dd}(\mathbf w_{d})_{k}\Bigg]\\
&=\mathbf{F}_{d-1}+\mathbf{f}_{d}.
\end{aligned}
\end{equation}
Appatently, the NS-NMF objective function in Equation \ref{eq:stream} is divided into two parts---$\mathbf{F}_{d-1}$ denotes the cost up to the $(d-1)$th sample and $\mathbf{f}_{d}$ denotes the cost of the $d$th sample, whereas $\mathbf{w}_{d}$ and $\mathbf{a}_{d}$ denote, respectively, the last row of $\mathbf{W}_{d}$ and $\mathbf{A}_{d}$. This strategy is known as the incremental NMF in the literature~\citep{sun:2018, chen:2018, zhang:2019b}. As the number of samples increases, new observations will have minor influence on the basis matrix, so that updating only the weight vector of the last sample will suffice.

In light of this idea, we can consider the first $d-1$ rows of $\mathbf{W}_{d}$ equal to $\mathbf{W}_{d-1}$. To compute the cost of $d$th sample, i.e., $\mathbf{f}_{d}$, we need to establish an updating policy for the basis matrix component $(\mathbf H_{d})_{kj}$ and weight matrix component $(\mathbf w_{d})_{k}$. We can utilize the gradient descent approach to derive the update policy as following:

\begin{equation}
\label{eq:stream1}
(\mathbf w_{d})_{k}^{t+1}=(\mathbf w_{d})_{k}^{t}-\delta_{k}\frac{\partial\mathbf{F}_{d}}{\partial(\mathbf w_{d})_{k}^{t}},
\end{equation}

\begin{equation}
\label{eq:stream2}
(\mathbf H_{d})_{kj}^{t+1}=(\mathbf H_{d})_{kj}^{t}-\theta_{kj}\frac{\partial\mathbf{F}_{d}}{\partial(\mathbf H_{d})_{kj}^{t}}.
\end{equation}
In Equations \ref{eq:stream1} and \ref{eq:stream2}, $\delta$ and $\theta$ denote the step sizes, $t$ denotes the iteration number, $k=1,2,\ldots,K$, and $j=1,2,\dots,p$. The step sizes are chosen as in Equations \ref{eq:stream3} and \ref{eq:stream35}, following the work in~\citet{cai:2011} and~\citet{guan:2012}, i.e.,

\begin{equation}
\label{eq:stream3}
\delta_{k}=\frac{(\mathbf w_{d})_{k}^{t}}{2(\alpha((\mathbf w_{d})^{t}\mathbf{H_{d}}^{t}(\mathbf{H_{d}}^{t})^T+\sum_{i=1}^{d}(\mathbf D_{d})_{id}\mathbf w_{i}^{t})_{k}},
\end{equation}

\begin{equation}
\label{eq:stream35}
\theta_{kj}=\frac{(\mathbf H_{d})_{kj}^{t}}{2\alpha(\mathbf W_{d-1}^T\mathbf W_{d-1}\mathbf H_{d}^{t}+(\mathbf w_{d}^{t+1})^T\mathbf w_{d}^{t+1}\mathbf H_{d}^{t})_{kj}}.
\end{equation}
Substituting Equations \ref{eq:stream3} and \ref{eq:stream35} into Equations \ref{eq:stream1} and \ref{eq:stream2}, respectively, we obtain the updating policy as follows:

\begin{equation}
\label{eq:stream4}
(\mathbf w_{d})_{k}^{t+1}=(\mathbf w_{d})_{k}^{t}\frac{(\alpha a_{d}(\mathbf{H}_{d}^{t})^T+\sum_{i=1}^{d-1}(\mathbf S_{d})_{id}\mathbf w_{i}^{t})_{k}}{(\alpha(\mathbf w_{d})^{t}\mathbf{H}_{d}^{t}(\mathbf{H}_{d}^{t})^T+(\mathbf D_{d})_{dd}\mathbf w_{d}^{t})_{k}},
\end{equation}

\begin{equation}
\label{eq:stream5}
(\mathbf H_{d})_{kj}^{t+1}=(\mathbf H_{d})_{kj}^{t}\frac{(\mathbf W_{d-1}^T\mathbf A_{d}+(\mathbf w_{d}^{t+1})^T\mathbf a_{d})_{kj}}{(\mathbf W_{d-1}^T\mathbf W_{d-1}\mathbf H_{d}^{t}+(\mathbf w_{d}^{t+1})^T\mathbf w_{d}^{t+1}\mathbf H_{d}^{t})_{kj}},
\end{equation}
where $\mathbf{S}_d$ and $\mathbf{D}_d$ represent, respectively, the MST-based weight matrix and degree matrix.

If we look at Equation \ref{eq:stream5}, it requires all the data samples before the $d$th one to compute the update. Doing so will obviously increase the memory requirements. To overcome the problem, we can use the strategy of cumulative summation. Let us first introduce two variables, $\mathbf{U}_{d}$ and $\mathbf{V}_{d}$, with the initial values of $\mathbf{U}_{0} = \mathbf{V}_{0}=\mathbf{0}$. Then using Equations \ref{eq:stream6} and \ref{eq:stream7} below, we can rewrite the update policy for $\mathbf{H}_{d}$ as in Equation \ref{eq:stream8} which now no longer needs to memorize all the samples.

\begin{equation}
\begin{aligned}
\label{eq:stream6}
\mathbf{U}_{d}&=\sum_{i=1}^{d}\mathbf{w}_{i}^{T}\mathbf{w}_{i}\\
&=\mathbf{U}_{d-1}+\mathbf{w}_{d}^{T}\mathbf{w}_{d}.
\end{aligned}
\end{equation}

\begin{equation}
\begin{aligned}
\label{eq:stream7}
\mathbf{V}_{d}&=\sum_{i=1}^{d}\mathbf{w}_{i}^{T}\mathbf{a}_{i}\\
&=\mathbf{V}_{d-1}+\mathbf{w}_{d}^{T}\mathbf{a}_{d}.
\end{aligned}
\end{equation}

\begin{equation}
\label{eq:stream8}
(\mathbf H_{d})_{kj}^{t+1}=(\mathbf H_{d})_{kj}^{t}\frac{(\mathbf V_{d})_{kj}}{\mathbf U_{d}(\mathbf H_{d}^{t})_{kj}}.
\end{equation}

Another problem with Equation \ref{eq:stream4} is that in order to update $(\mathbf w_{d})_{k}$, we need to reconstruct the MST with all the data samples every time as a new sample comes in. Again, doing so slows down the online operation.  To address this problem we adopt the combination of local MST~\citep{ahmed:2019a} and buffering strategy~\citep{goldberg:2008, liu:2016}. In a local MST, for each observation we construct a MST with its neighbors only. These neighbors can be chosen in a temporal fashion. In other words, the neighbors of the $d$th sample (which just arrives) are the samples having arrived in a specified time window before it, say, in the window of dating back to time instance $d-z$, where $z$ is the size of the time window. If $z$ =50, it means that the MST will be constructed based on the $d$th sample and the most recent 50 samples arrived before the $d$th sample. The buffering strategy states that instead of discarding all the old data samples, one maintains a buffer of specified size, $Q$. After the buffer is full for the first time, any new sample will be added to the buffer while the buffer drops the oldest one, thereby keeping its size the same. To connect the two approaches, we set the time window size the same as the buffer size, i.e., $Q=z$. Consequently, we can rewrite the updating policy for $(\mathbf w_{d})_{k}$ as

\begin{equation}
\label{eq:stream9}
(\mathbf w_{d})_{k}^{t+1}=(\mathbf w_{d})_{k}^{t}\frac{(\alpha \mathbf a_{d}(\mathbf{H}_{d}^{t})^T+\sum_{i=d-z}^{d-1}(\mathbf S_{d})_{id}\mathbf w_{i}^{t})_{k}}{(\alpha(\mathbf w_{d})^{t}\mathbf{H}_{d}^{t}(\mathbf{H}_{d}^{t})^T+(\mathbf D_{d})_{dd}\mathbf w_{d}^{t})_{k}}.
\end{equation}

The algorithm steps are summarized in Algorithm~\ref{algorithm2}. There are three phases of the algorithm after the initialization. Steps 5--8 summarizes the first phase, i.e., $d < z$, where the new samples are added along with initialization of the weight vectors. When the buffer is full for the first time, i.e., $d = z$, it starts the second phase, in which an MST is constructed and the offline NS-NMF algorithm helps obtain the weights and basis vectors. Step 9--13 summarizes this phase. After that, the algorithm enters the third and final phase, i.e., $d > z$, where the update of the weight and basis vectors and calculation of the anomaly scores are carried out in Steps 14--23. In this phase, a gradient descent approach is used to obtain the updated weight and basis values for each new sample. Steps 16--20 summarizes the iterations on $t$, which are required for the convergence of the gradient descent approach.

\begin{algorithm}[tb]
\caption{Online implementation of NS-NMF algorithm for anomaly detection}
\label{algorithm2}
    \SetKwInOut{Input}{Input}
    \SetKwInOut{Output}{Output}
    \Input{Current observation, $\mathbf{a}_{d}$, trade-off parameter, $\alpha$, buffer size, $z$, and the number of clusters, $K$}
    \Output{Current
 basis matrix, $\mathbf{H}_{d}$, and anomaly score, $O_{d}$}
     Initialize $\mathbf{H}_{0}$ with random values\;
     Initialize $\mathbf{U}_{0} = \mathbf{V}_{0} =\mathbf{0}$\;
     Initialize $\mathbf{A}_{0} = \mathbf{W}_{0} =\mathbf{S}_{0}= \phi$\;
    \While{a new observation $\mathbf{a}_{d}$ arrives} {
        Draw the current sample $\mathbf{a}_{d}$\;
        Initialize weight coefficient $\mathbf{w}_{d}$ with random values\;
        Append $\mathbf{a}_{d}$ to $\mathbf{A}_{d-1}$ and assign it to $\mathbf{A}_{d}$\;
        Append $\mathbf{w}_{d}$ to $\mathbf{W}_{d-1}$ and assign it to $\mathbf{W}_{d}$\;
        \If{$d=z$}{
            Construct MST using the observations in $\mathbf{A}_{d}$ and store the weights in $\mathbf{S}_{d}$\;
            Apply offline NS-NMF to obtain $\mathbf{W}_{d}$ and $\mathbf{H}_{d}$\;
            Calculate $\mathbf{U}_{d}$ and $\mathbf{V}_{d}$ using Equations \ref{eq:stream6} and \ref{eq:stream7}\;
            }
        \If{$d>z$}{
             Construct MST using the observations in $\mathbf{A}_{d}$ and obtain $\mathbf{s}_{d}$\;
            \Repeat{Convergence}{
               Use  Equation \ref{eq:stream9} to update $\mathbf{w}_{d}^{t+1}$ using $\mathbf{s}_{d}$ and $\mathbf{H}_{d}$\;
               Use Equations \ref{eq:stream6} and \ref{eq:stream7} to update $\mathbf{U}_{d}$ and $\mathbf{V}_{d}$ with $\mathbf{w}_{d}^{t+1}$\;
               Use  Equations \ref{eq:stream9} to update $\mathbf{H}_{d}^{t+1}$\;
              }

           Delete the first row vector from both $\mathbf{A}_{d}$ and $\mathbf{W}_{d}$\;
           Calculate the anomaly score for the $d$th observation using Equation \ref{eq:stream10} and store it in $O_{d}$
            }
     }
\end{algorithm}

Similar to the offline counterpart, we can compute the anomaly score of the $d$th observation as in Equation \ref{eq:stream10}.
\begin{equation}
\label{eq:stream10}
O_{d} = \left \|\mathbf {a}_{d}-{\mathbf {a}_{d}}'  \right \|_{2} = \left \|\mathbf {a}_{d}-{\sum_{k=1}^K (\mathbf{w}_{d})_{k} (\mathbf{H}_{d})_{k}}  \right \|_{2}.
\end{equation}
Now, we can either choose a threshold and mark the observation as anomaly on the fly if its anomaly score crosses the threshold or we can store the anomaly scores to do the evaluation later. In this work, we test both of the options. For the benchmark data sets, which do not have any timestamps marking, we decide to go for the second option. What this means is that while we run the algorithm to get the anomaly scores as the algorithm sieves through the data sequence, the declaration of anomaly is based on selecting the top $N$ scores as anomalies, the same as we do in the offline scenario. For the hydropower data set, on the other hand, we do have the associated timestamps, so we decide to go with the first option and mark anomalies on the fly.

\section{Comparative Performance Analysis of NS-NMF}\label{sec:performance}

We evaluate the performance of the proposed NS-NMF method for anomaly detection, as compared to the vanilla NMF/GNMF/SNMF (in Section~\ref{section5.1}) and to a non-NMF method (in Section~\ref{section5.1b}). In Sections~\ref{section5.1} and~\ref{section5.1b}, we use 20 benchmark anomaly detection data sets from the study of \cite{campos:2016} for our performance comparison study. In Table~\ref{tbl:1}, we summarize the basic characteristics of these 20 data sets. For all the benchmark data sets, the label of the observations whether it is normal or anomalous is known beforehand. There are several versions of these data sets available depending on the data cleaning and preprocessing steps involved. For our analysis we choose to use the normalized version of the data sets with all missing values removed and categorical variables are converted into numerical format. In Section~\ref{section5.2}, we apply the competitive methods to a real-life data set from a hydropower plant.

\begin{table}[tb]
\centering
\caption{Public benchmark data sets used in the performance evaluation study.}
\label{tbl:1}

\begin{tabular}{|l|l|l|l|l|l|l|l|}
\hline
\multirow {2}{*}{Data set}          & Number of  & Number of  & Number of \\
          & observations ($n$)      & anomalies ($|O|)$ & attributes ($p$)\\
\hline
Annthyroid       & 7,200   & 347  & 21
\\ \hline
Arrhythmia       & 450    & 12   & 259
\\ \hline
Cardiotocography & 2,126   & 86   & 21
\\ \hline
Heart     & 270    & 7    & 13
\\ \hline
Page Blocks      & 5,473   & 99   & 10
\\ \hline
Parkinson        & 195    & 5    & 22
\\ \hline
PIMA             & 768    & 26   & 8
\\ \hline
SpamBase         & 4,601   & 280  & 57
\\ \hline
Stamps           & 340    & 16   & 9
\\ \hline
WBC              & 454    & 10   & 9
\\ \hline
Waveform         & 3,443   & 100  & 21
\\ \hline
WPBC             & 198    & 47   & 33
\\ \hline
WDBC             & 367    & 10   & 30
\\ \hline
ALOI             & 50,000  & 1,508 & 27
\\ \hline
KDD         & 60,632 & 200  & 41
\\ \hline
Shuttle          & 1,013   & 13   & 9
\\ \hline
Ionosphere       & 351    & 126  & 32
\\ \hline
Glass            & 214    & 9    & 7
\\ \hline
Pen digits       & 9,868   & 20   & 16
\\ \hline
Lymphography     & 148    & 6    & 19
\\ \hline
\end{tabular}
\end{table}

To evaluate the performance of the methods, the criterion we use is called \textit{precision at $N$}\citep[$P@N$]{campos:2016}, which is a rather common performance metric used in anomaly detection. As mentioned earlier, in a practical setting for anomaly detection, researchers often set a cut-off threshold $N$ and flag the observations with the top-$N$ anomaly scores, however it is defined and computed in respective methods.  Ideally, $N$ is chosen to be the number of true anomalies.  Practically, when the number of true anomalies is unknown, $N$ is chosen to be larger than the perceived number of anomalies but small enough to make the subsequent identification operations feasible.

In the benchmark study, since we know the number of true anomalies,  we therefore use that value as our choice of $N$ and treat it as the same cut-off for all methods in comparison.  Because $N$ is the number of true anomalies, the number of false alarms is implied, which is $N-N\times P@N$.  That is why in the benchmark study, we only present $P@N$. In reality, when the number of true anomalies is not known, the main objective in anomaly detection is still to increase $P@N$ for a fixed $N$, i.e., to have a higher detection rate within the cut-off threshold.

The $P@N$ is defined as the proportion of correct anomalies identified in the top $N$ ranks. For a data set $DB$ of size $n$, consisting of anomaly set $O$ $\subset$ $DB$ and normal data sets $I$ $\subseteq$ $DB$, such that $DB$ = $O$ $\cup$ $I$, $P@N$ can be formulated as
\begin{equation} \label{eq:evaluation}
P@N = \frac{\#\{o  \in O \, | \, rank(o) \le N \}}{N},  \quad \text{where} \quad N = |O|.
\end{equation}

\subsection{Comparison with Other NMF Methods}\label{section5.1}
First, let us take a look at the parameter selection policies for the competing methods within the NMF framework. The number of latent features or clusters, $K$, is needed for all of the NMF-based methods. In this study we use $K =5$. We have also explored the possibility of using $K=2, 10, 15, 20$ and $25$ and found that changing $K$ in this range does not affect NMF-methods a great deal. More importantly, the relative performance among the competing NMF methods remain the same. Based on our experiments with different $K$'s, we observe that the NMF-based anomaly detection methods perform better when $K$ is in the range of $[5, 15]$. We settle for $K=5$ because it results in overall good detection performances for all competing methods. The cut-off value, $N$, required to generate the final anomaly list for both offline and online version, is taken as the number of true anomalies in the benchmark data set studies, as discussed above. Other than these two parameters that are common to all methods, the rest of the parameters are algorithm-specific. We summarize the parameter choices in Table~\ref{tbl:2}. For SNMF and GNMF, we adopt the best settings described in their original papers. For NS-NMF, as mentioned earlier, its performance is not sensitive to the choice of $\alpha$.  We settle at $\alpha=0.8$ by conducting a few simple trials.

\begin{table}[]
\centering
\caption{Parameter values and settings used for NS-NMF, GNMF and SNMF.}
\label{tbl:2}
\begin{tabular}{|l|l|l|l|l|}
\hline
\small
Competing & Number of  & \multirow {2}{*}{Other settings}\\
methods & latent factors, $K$ & \\\hline
\multirow {2}{*}{Offline NS-NMF} & \multirow {2}{*}{5}  & Parameter controlling the influence of NMF, $\alpha = 0.8$ \\
                                                                        & & Regularizer for controlling overfitting, $\gamma = 0.2$\\ \hline
\multirow {2}{*}{Online NS-NMF} & \multirow {2}{*}{5}  & Parameter controlling the influence of NMF, $\alpha = 0.8$ \\
                                                                        & & Buffer size, $z=B = 20$\\ \hline
\multirow {3}{*}{GNMF}& \multirow {3}{*}{5} & Manifold regularizer, $\lambda = 100$ \\
          & &Neighborhood graph construction parameter, $q = 5$\\
          & &Weighting scheme for adjacency matrix: $0-1$\\\hline
SNMF              & 5                          & Gaussian similarity measure for constructing $\mathbf{S}$                   \\ \hline
\end{tabular}

\end{table}

\begin{figure}[p]
	\centering
	\centerline{\includegraphics[width=0.9\textwidth]{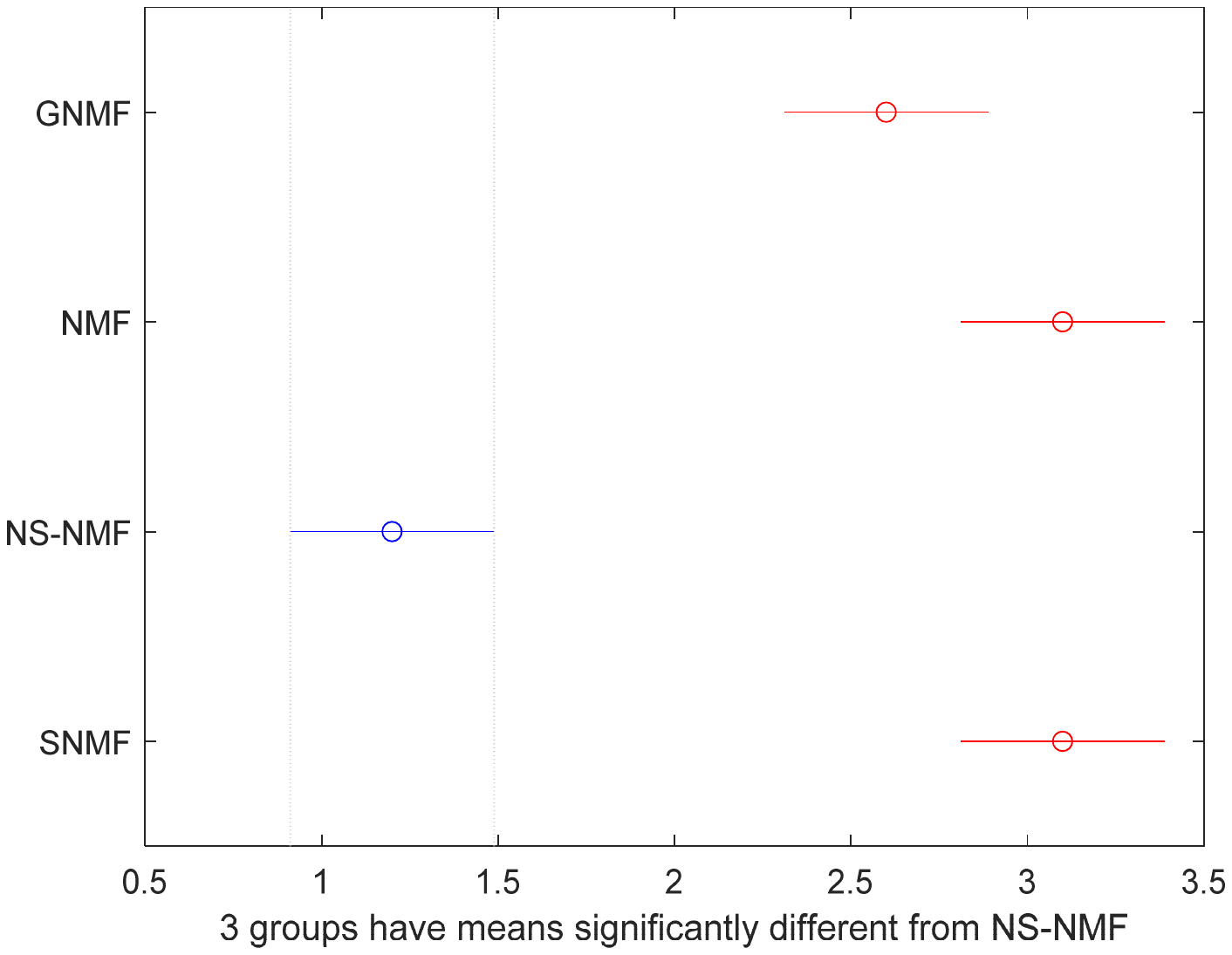}}
\caption{Post hoc analysis on the ranking data obtained by the Friedman test.}
	\label{fig:posthoc}
\end{figure}

We present the comparison of detection performance of the four offline methods on the 20 benchmark data sets in Table~\ref{tbl:3}. For this comparison, we only consider the offline version of NS-NMF because the competitors are offline, so it is a bit unfair if we compare the online version of NS-NMF. For this reason, NS-NMF means the offline NS-NMF in the comparison shown in Table~\ref{tbl:3}, Table~\ref{tbl:11}, and Figure~\ref{fig:posthoc}.

To better reflect the methods' comparative edge, we break down the comparison into four major categories in Table~\ref{tbl:3}, namely \textit{Better}, \textit{Equal}, \textit{Close} and \textit{Worse}, as explained in the table. NS-NMF outperforms other methods by showing uniquely best detection performance on 16 out of 20 data sets and tying for the best in another three cases.  Only in one case NS-NMF is obviously worse than the best performer. The vanilla NMF achieves the uniquely best performance in a single case, while GNMF and SNMF tie for some best performance but never beat others outright. If we rank each of these four methods in a scale of 1 (best) to 4 (worst), then the mean rank for NS-NMF is 1.2, which is far ahead of other methods, with GNMF at 2.6 mean rank, SNMF and NMF both at 3.1 mean rank.

\begin{table*}[tb]
\centering
\caption{Performance comparison.}
\label{tbl:3}
\begin{tabular}{|l|l|l|l|l|l|l|l|l|l|l|l|l|l|l|l|l|}
\hline
\diaghead(5,-1){PerformanceOutlier detection methodsxxxx}{Performance (number of data sets)}{Anomaly detection methods}& NS-NMF & NMF & GNMF & SNMF \\ \hline
Better (uniquely best result)        & 16   & 1   & 0   & 0    \\\hline
Equal (equal to the existing best result)            & 3   & 2   & 2   & 2     \\\hline
Close (within 20\% of the best result)     & 0  & 4   & 9   & 6      \\\hline
Worse (not within 20\% of the best result) & 1   & 13   & 9   & 12    \\ \hline
Mean relative rank        & 1.2   & 3.1  & 2.6  & 3.1      \\ \hline
\end{tabular}

\end{table*}

We apply the Friedman test, a non-parametric method \citep{demšar:2006}, to determine whether NS-NMF achieves significant improvement over other competitors. Let $n_{a}$ be the number of anomaly detection methods and $n_{d}$ be the number of data sets. We define a matrix $Ra$ whose entries in each row represent the detection method's rank for that specific data set. If there are tied values, we assign to each tied value the average of the ranks that would have been assigned without ties. For example, suppose we have two tied methods both with rank 1. Had there been no tie, one should have been assigned rank 1 and the other rank 2. An Friedman test then uses the average of the two ranks, which is 1.5, as the rank value for both of these methods. Under the null hypothesis that all methods perform the same, the Friedman statistic,
\begin{equation} \label{eq:frd}
\chi^{2}_{F} = \frac{12n_{d}}{n_{a}(n_{a}+1)} \left(\sum_{l=1}^{n_{a}}\overline{Ra_{l}^{2}}-\frac{n_{a}(n_{a}+1)^{2}}{4}\right),
\end{equation}
follows a Chi-squared distribution with $n_{a}-1$ degrees of freedom, where $\overline{Ra_{l}}$ is the average value of column $l = 1, 2, \ldots, n_{a}$. We  found the p-value ($1.23\times10^{-7}$) significant enough to reject the null hypothesis.

We also perform a post hoc analysis of the four methods in terms of their ranking performance. Fig. \ref{fig:posthoc} presents the post hoc analysis on the ranking data and it indicates that the ranking of NS-NMF is significantly better than the other three approaches at the 0.01 level of significance. The detailed pairwise comparisons between NS-NMF and each of the three methods are presented in Table~\ref{tbl:11}. The p-values are calculated using the Conover post-hoc test~\citep{conover:1999}. We employ the Bonferroni correction~\citep{bland:1995} to adjust the p-values for multiple comparisons. All the pairwise comparisons show a sufficiently significant difference.

\begin{table}[tb]
\centering
\caption{The p-values of pairwise comparisons between the NS-NMF method with each of the three competing methods.}
\label{tbl:11}

\begin{tabular}{|l|l|l|}
\hline
 Competing  methods     & p-value    \\ \hline
NS-NMF vs NMF   & $7.11\times 10^{-21}$           \\ \hline
NS-NMF vs GNMF   & $3.40\times 10^{-15}$        \\ \hline
NS-NMF vs SNMF   & $7.11\times 10^{-21}$     \\ \hline
\end{tabular}

\end{table}

In Table~\ref{tbl:TPR}, we summarize the number of true detections by the competing methods.  We notice that the offline NS-NMF either outperforms or is no worse than both GNMF and SNMF in every single case. GNMF's performance is better than that of NMF and SNMF, although still overall worse than the offline NS-NMF. We use the best parameter setting for GNMF as recommended by its authors.  We do observe the sensitive nature of GNMF to its parameters and acknowledge the possibility that some other parameter combinations might produce a better outcome. However, parameter selection in unsupervised learning settings is a difficult task, as the common approaches working well for supervised learning like cross validation does not work in the unsupervised circumstances.  Therefore, the sensitivity of GNMF is certainly a shortcoming.  SNMF's performance is surprisingly not up to the mark and produces worse results than the vanilla NMF in seven cases. SNMF is also the slowest in terms of computational time and it does not appear scalable on big data sets.  For example, in the case of the ``ALOI'' data set that has 50,000 observations, SNMF takes almost eight hours to generate the index of the anomalous observations, whereas offline NS-NMF takes one quarter of that time to complete the task, with the majority of its time spent on MST construction. NMF and GNMF are much faster and take around 12 mins and 35 mins, respectively, in processing the same data set.

\begin{table*}[tb]
\centering
\caption{Number of true positive detections of the competing methods. Bold entries represent the best detection performance in a respective data set.}
\label{tbl:TPR}
\begin{tabular}{|l|l|l|l|l|l|l|l|l|l|l|l|l|l|l|}
\hline
\multirow {2}{*}{Data set}   & Offline  & Online & NMF & GNMF & SNMF & Total \\
        &NS-NMF &NS-NMF & & & &Anomalies \\
\hline
Glass             & \textbf{4}              & 2             & 1   & 1    & 1    & 9               \\ \hline
PIMA              & \textbf{8}            & 4             & 2   & 2    & 2    & 26              \\ \hline
WBC               & \textbf{9}             & 7             & 2   & 8    & 8    & 10              \\ \hline
Stamps            & \textbf{4}              & 2             & 3   & 3    & 2    & 16              \\ \hline
Shuttle           & \textbf{2}              & \textbf{2}             & 0   & 0    & 0    & 13              \\ \hline
Pageblocks        & 21             & 27            & \textbf{30}  & 19   & 14   & 99              \\ \hline
Heart             & \textbf{5}              & 4             & 0   & 4    & 4    & 7               \\ \hline
Pendigits         & 0              & \textbf{1}             & 0   & 0    & 0    & 20              \\ \hline
Lymphography      & \textbf{5}              & 4             & 4   & 4    & 2    & 6               \\ \hline
Waveform          & \textbf{9}              & \textbf{9}             & 3   & 6    & 4    & 100             \\ \hline
Cardiotocograpghy & \textbf{29}             & 15            & 27  & \textbf{29}   & \textbf{29}   & 86              \\ \hline
Annthyroid        & \textbf{27}             & 15            & 15  & 13   & 10   & 347             \\ \hline
Parkinson         & \textbf{3}              & \textbf{3}             & 1   & 2    & 2    & 5               \\ \hline
ALOI              & \textbf{193}            & 145           & 78  & 117  & 87   & 1508            \\ \hline
WDBC              & \textbf{7}              & 6             & 1   & 6    & 6    & 10              \\ \hline
Ionosphere        & \textbf{92}             & 74            & 79  & 74   & 33   & 126             \\ \hline
WPBC              & \textbf{11}             & 9             & \textbf{11}  & 8    & 9    & 47              \\ \hline
KDD               & \textbf{102}            & 97            & 49  & 93   & 75   & 200             \\ \hline
Spambase          & \textbf{47}             & 41            & 32  & 36   & 23   & 280             \\ \hline
Arrhythmia        & \textbf{4}              & 3             & 2   & 3    & 3    & 12              \\ \hline
\end{tabular}

\end{table*}

Here we also include the results from the online version of NS-NMF because we would like to draw a comparison between the offline and online NS-NMF and see how much efficiency the online NMF maintains while using a small subset of data to compute the anomaly scores. Unsurprisingly, the offline version comes out superior in 15 out of the 20 cases. On the other hand, for most of the cases, the online version does not perform that much worse than the offline version. A bit surprisingly, the online version even beats its offline counterpart in two cases. It seems to suggest that in some cases, having a longer memory and global data connection may not always help the detection. It is also interesting to note that online NS-NMF produces comparatively better results than other NMF variants, excluding offline NS-NMF.  For instance, once excluding offline NS-NMF, online NS-NMF attains the best detection 14 times (some are ties), whereas GNMF seven times, SNMF five times, and vanilla NMF six times. This detection performance is remarkable because the main advantage of online NS-NMF lies in its considerable improvement in run time. For example, for the ``ALOI'' data set, which has 50,000 data points, online NS-NMF achieves an approximately 32-fold improvement over SNMF in terms of time expense. Overall we consider the online NS-NMF method a competent and promising online anomaly detection algorithm.

\subsection{Comparison with non-NMF Methods}\label{section5.1b}
As NS-NMF outperforms other NMF-based approaches quite comprehensively, we would also like to see how it performs compared to non-NMF anomaly detection approaches. We choose one particular non-NMF method as the representative in this comparison, which is known as the Local MST~\citep[LoMST]{ahmed:2019a}. There are two reasons behind our choice here: (1) LoMST has gone through a comparison study involving 14 methods over 20 data sets and emerged as the winner. (2) Comparing with NS-NMF, LoMST employs the MST mechanism to form its similarity measure but does not go through a dimensionality reduction process, whereas NS-NMF integrates the MST similarity measure while it creates the low-dimensional embedding.  The comparison in Section~\ref{section5.1} establishes that having MST almost always helps anomaly detection within the NMF framework.  The question is then whether using MST without NMF would be equally effective.  In other words, does one really need NMF in the first place or can one simply opt to do an MST-based detection on the original data?

The parameter selection process for NS-NMF remains the same as in Section~\ref{section5.1}.  The LoMST approach requires the number of the nearest neighbors ($k$) as a user input. In~\citet{ahmed:2019a}, two scenarios of how $k$ is decided were discussed---the best $k$ scenario, in which the optimal $k$ is selected after many choices of $k$ were tried and the detection outcomes were compared with the ground truth (recall the ground truth for the benchmark data sets is known), and the practical $k$ scenario, which is to choose the $k$ value that returns the maximum standard deviation of the LoMST scores, without knowing the detection ground truth.  In reality, the best $k$ scenario is impractical.  So in this comparison, we use the practical $k$ for LoMST. We also want to note a difference in terms of data use in this comparison versus that in~\citet{ahmed:2019a}.  When we apply the NS-NMF method (in fact, any of the NMF variants) to the Waveform data set, because some of its entries are negative, we use the normalized data set version, which scales the data to the range of $[0,1]$.  This normalized version is what we continue using in this comparison, while the Waveform data set used in~\citet{ahmed:2019a} was without the normalization.

The comparative performance of NS-NMF and LoMST is listed in Table~\ref{tbl:TPRL}. From the table, we see that NS-NMF overall outperforms LoMST. In 14 cases out of 20, NS-NMF turns out to be a winner, whereas LoMST produces a better detection performance in five cases.  The two methods tie in one case.

\begin{table*}[h]
\centering
\caption{Comparison with the LoMST approach. Bold entries represent the best detection performance in a respective data set.}
\label{tbl:TPRL}
\begin{tabular}{|l|l|l|l|l|l|}
\hline
Data set           & NS-NMF & LoMST  & Number of attributes  \\ \hline
Glass             & \textbf{4}      & 3      & 7              \\ \hline
PIMA              & \textbf{8}     & \textbf{8}       & 8              \\ \hline
WBC               & \textbf{9}      & 8      & 9              \\ \hline
Stamps            & \textbf{4}      & 2      & 9              \\ \hline
Shuttle           & 2      & \textbf{6}       & 9              \\ \hline
Pageblocks        & 21     & \textbf{32}      & 10              \\ \hline
Heart             & \textbf{5}      & 4      & 13             \\ \hline
Pendigits         & 0      & \textbf{2}       & 16              \\ \hline
Lymphography      & 5      & \textbf{6}       & 19              \\ \hline
Waveform          & \textbf{9}      & 1      & 21             \\ \hline
Cardiotocograpghy & \textbf{29}     & 27    & 21              \\ \hline
Annthyroid        & \textbf{27}     & 16     & 21             \\ \hline
Parkinson         & \textbf{3}      & 2      & 22               \\ \hline
ALOI              & \textbf{193}    & 185    & 27            \\ \hline
WDBC              &\textbf{7}      & 6      & 30              \\ \hline
Ionosphere        & 92     & \textbf{101}    & 32             \\ \hline
WPBC              & \textbf{11}     & 9      & 33              \\ \hline
KDD               & \textbf{102}    & 39     & 41             \\ \hline
Spambase          & \textbf{47}     & 45    & 57             \\ \hline
Arrhythmia        & \textbf{4}      & 3      & 259              \\ \hline
\end{tabular}
\end{table*}

If we further analyze the result, we see that NS-NMF performs consistently better than LoMST when the number of attributes (4th column in Table~\ref{tbl:TPRL}) is relatively large. For the 20 data sets, we can partition them into two roughly equal halves, where the upper half has a relatively smaller number of attributes and the bottom half has relatively larger attributes. The threshold used for partitioning the number of attributes is 20. Nine data sets are in the upper half and 11 data sets are in the bottom half.  Out of the 11 data sets in the bottom half, NS-NMF won 10 cases and lost one, whereas for the nine cases in the upper half, NS-NMF won four cases, LoMST won four cases, and they tied for one case

We believe that this pattern did not happen entirely by chance but it is related to the anomaly detection mechanisms used by the two competing approaches. NS-NMF, because of the inclusion of NMF, performs dimensionality reduction before it attempts to detect anomalies, whereas LoMST, on the other hand, is a neighborhood-based method that detects anomalies in the original space without dimensionality reduction. The difference between NS-NMF and LoMST is really due to the need and benefit of dimensionality reduction. While researchers consider dimensionality reduction generally helpful, there is no guarantee that it always helps in every single case.  It makes sense that the need of dimensionality reduction is more acute and the benefit would be more likely and pronounced when the data spaces are of higher dimensions.  Our empirical analysis above appears to be consistent with these general understandings.

\subsection{Application to Power Plant Data}\label{section5.2}

The industry data set used in this study comes from a hydropower plant. The same data set has been analyzed in our previous work~\citep{ahmed:2019a, ahmed:2018}. To quickly recap the basic information of the data set, we have a total of seven months worth of data, coming from different functional areas of the plant (turbines, generators, bearings etc.). We combine all the data from different functional areas according to their time stamps and perform some simple cleaning, statistical analysis and pre-processing; additional details about the data pre-processing can be found in~\cite{ahmed:2018}. In the end, we have $n=9,219$ observations and $p=222$ attributes. Then we apply NS-NMF, GNMF, SNMF and the vanilla NMF algorithms to find out the anomalies in this data set. We use the same parameter settings as we do in the benchmark data set study. The only difference is that we no longer have the information of the number of true anomalies. After consulting the operating manager who provided this data set, we are advised to report top 100 anomalous time stamps to the manager. The manager would follow up and check the status of operation in detail, for instance, by manually examining operational logs and physical conditions of components, and then confirm us about the validity of the findings.

\begin{table}[p]
\centering
\caption{Summary of the top 100 anomalies.}
\label{tbl:8}
\scalebox{0.9}{
\begin{tabular}{|c|c|c|c|c|}
\hline
Offline NS-NMF  & Online NS-NMF         & GNMF             & SNMF           & NMF  \\ \hline
4/16/2015 2:40    & 7/4/2015 12:10 & 7/4/2015 11:30   & 4/15/2015 16:50 & 7/4/2015 11:30  \\ \hline
4/16/2015 4:30    & 7/4/2015 12:20 & 7/4/2015 11:40   & 4/15/2015 18:30 & 7/4/2015 11:40 \\ \hline
7/4/2015 11:30   & 7/4/2015 12:30 & 7/4/2015 11:50   & 4/15/2015 20:10 & 7/4/2015 11:50 \\ \hline
7/4/2015 11:40   & 7/4/2015 12:40 & 7/4/2015 12:00   & 4/15/2015 22:30 & 7/4/2015 12:00  \\ \hline
7/4/2015 11:50   & 7/4/2015 12:50 & 7/4/2015 12:10   & 4/16/2015 2:50 & 7/4/2015 12:10 \\ \hline
7/4/2015 12:00   & 7/4/2015 1:00 & 7/4/2015 12:20   & 4/16/2015 4:30 & 7/4/2015 12:20   \\ \hline
7/4/2015 12:10   & 7/4/2015 1:20 & 7/4/2015 12:30   & 4/16/2015 4:50 & 7/4/2015 12:30   \\ \hline

$..................$         & $..................$ & $..................$  &  $..................$   &  $..................$      \\ \hline
9/13/2015 19:00 & 9/14/2015 7:30 & 7/7/2015 4:20   & 4/19/2015 3:40 & 7/4/2015 16:20 \\ \hline
9/13/2015 19:10   & 9/14/2015 7:40 & 7/8/2015 12:20   & 4/19/2015 4:00 & 7/4/2015 16:30  \\ \hline
9/14/2015 8:00   & 9/14/2015 7:50 & 7/8/2015 18:00   & 4/19/2015 21:40 & 7/4/2015 16:40 \\ \hline
9/14/2015 8:10   & 9/14/2015 8:00 & 9/15/2015 21:20  & 4/19/2015 23:40 & 7/4/2015 16:50 \\ \hline
9/14/2015 8:20   & 9/14/2015 8:20 & 9/15/2015 21:40  & 4/20/2015 8:20  & 7/4/2015 17:00 \\ \hline
9/14/2015 8:30   & 9/14/2015 8:50 & 10/3/2015 18:50  & 7/4/2015 11:20  & 7/4/2015 17:20 \\ \hline
9/14/2015 8:40  & 9/14/2015 13:00 & 10/3/2015 19:10  & 7/4/2015 11:30 & 7/4/2015 17:30  \\ \hline
9/14/2015 13:00  & 9/14/2015 13:10 & 10/3/2015 19:20  & 7/4/2015 11:50 & 7/4/2015 17:40   \\ \hline

$..................$      & $..................$    & $..................$  &  $..................$  &  $..................$       \\ \hline
10/3/2015 21:00  & 10/3/2015 20:10 & 10/3/2015 22:20  & 7/4/2015 15:10  & 10/3/2015 18:20 \\ \hline
10/3/2015 21:10  & 10/3/2015 20:20 & 10/3/2015 22:30  & 7/4/2015 15:20 & 10/3/2015 18:50  \\ \hline
10/3/2015 21:20  & 10/3/2015 20:30 & 10/3/2015 22:40  & 7/4/2015 15:30 & 10/3/2015 19:10  \\ \hline
10/3/2015 21:30  & 10/3/2015 20:50 & 10/3/2015 22:50  & 7/4/2015 15:40 & 10/3/2015 19:20  \\ \hline
10/3/2015 21:40  & 10/3/2015 21:00 & 10/3/2015 23:00  & 7/4/2015 15:50 & 10/3/2015 19:30  \\ \hline
10/3/2015 21:50  & 10/3/2015 21:10 & 10/3/2015 23:10  & 7/4/2015 16:00 & 10/3/2015 19:40  \\ \hline
10/3/2015 22:10  & 10/3/2015 21:20 & 10/3/2015 23:20  & 7/4/2015 16:20 & 10/3/2015 19:50  \\ \hline

$..................$       & $..................$   & $..................$  &  $..................$ &  $..................$        \\ \hline
10/4/2015 0:10   & 10/3/2015 23:30 & 10/4/2015 17:40  & 7/4/2015 18:10 & 10/3/2015 23:10   \\ \hline
10/4/2015 0:20   & 10/3/2015 23:50 & 10/4/2015 17:50  & 7/4/2015 18:20 & 10/3/2015 23:20  \\ \hline
10/4/2015 0:30   & 10/4/2015 0:00 & 10/4/2015 18:00  & 7/8/2015 15:50 & 10/3/2015 23:30   \\ \hline
10/4/2015 23:40  & 10/4/2015 0:10 & 10/4/2 015 18:10  & 7/8/2015 18:00 & 10/4/2015 0:00  \\ \hline
10/4/2015 23:50  & 10/4/2015 0:20 & 10/4/2015 18:20  & 9/17/2015 4:30 & 10/4/2015 0:10  \\ \hline
10/5/2015 1:00   & 10/4/2015 0:30 & 10/4/20 15 18:30  & 9/17/2015 4:40 & 10/4/2015 0:20  \\ \hline
10/5/2015 1:30   & 10/4/2015 22:50 & 10/4/2015 18:40  & 9/17/2015 4:50 & 10/4/2015 0:30  \\ \hline

$..................$        & $..................$  & $..................$  &  $..................$   &  $..................$      \\ \hline
10/5/2015 4:40   & 10/13/2015 17:25 & 10/5/2015 3:50   & 10/3/2015 19:10 &10/5/2015 2:30  \\ \hline
10/5/2015 4:50   & 10/13/2015 17:30 & 10/5/2015 4:00   & 10/3/2015 19:20 &10/5/2015 3:00 \\ \hline
10/13/2015 16:25   & 10/13/2015 17:35 & 10/5/2015 4:10   & 10/3/2015 19:30  & 10/5/2015 3:10\\ \hline
10/13/2015 16:35 & 10/13/2015 17:45 & 10/5/2015 4:20   & 10/3/2015 19:40  & 10/5/2015 3:20\\ \hline
10/13/2015 16:45 & 10/13/2015 18:20 & 10/5/2 015 4:30   & 10/3/2015 19:50  & 10/5/2015 3:30\\ \hline
10/13/2015 16:55 & 10/13/2015 18:30 & 10/5/2015 4:50   & 10/3/2015 20:40  & 10/5/2015 3:40\\ \hline

$..................$       & $..................$   & $..................$  &  $..................$   &  $..................$      \\ \hline
1/2/2016 21:20   & 1/11/2016 18:10 & 10/13/2015 17:05 & 10/13/2015 17:05 & 10/13/2015 17:25\\ \hline
1/2/2016 21:30   & 1/12/2016 11:20 & 10/13/2015 17:15 & 10/13/2015 17:15 & 10/13/2015 17:35\\ \hline
1/2/2016 21:40   & 1/12/2016 11:30 & 10/13/2015 17:25 & 10/13/2015 17:25 & 10/13/2015 17:45\\ \hline
1/12/2016 11:20  & 1/12/2016 11:40 & 10/13/2015 17:35 & 10/13/2015 17:35 & 10/13/2015 17:55\\ \hline
1/12/2016 11:30  & 1/12/2016 12:10 & 10/13/2015 17:45 & 10/13/2015 17:45   & 10/13/2015 18:20 \\ \hline

\end{tabular}}
\end{table}

The top 100 anomalies identified by the four methods are shown in Table \ref{tbl:8}. We use both online and offline version of NS-NMF. To detect anomalies on the fly, we use the threshold update policy similar to that in~\citet{ahmed:2019b} for the online algorithm. To save space we skip some rows in the table. We observe that altogether these four methods have 27 common anomalous time stamps among the top 100 anomalies, whereas offline NS-NMF and online NS-NMF produce similar outcomes and share 46 common time stamps. These common findings serve as an indirect way of cross validating the sanity of the detection outcomes.

We find that the anomalous time stamps belong to a few anomaly-prone days, which are listed in Table \ref{tbl:9}. It is also noticeable that anomalies occur in chunks, and in most cases, the observations in the close time vicinity of an anomalous time stamp are also returned as anomalies. When we report these time stamps to the operating manager, he agrees, after his own verification, that most of these findings present valid concerns and indeed require trouble shooting.

While there is a good common ground shared by these methods, the competing methods do perform differently at certain aspect. GNMF entirely misses the anomalous stamps on September 13\textsuperscript{th}, September 14\textsuperscript{th}, January 2\textsuperscript{nd}, and January 12\textsuperscript{th}. SNMF likewise misses those dates. The offline NS-NMF successfully identified all of those time stamps, but misses some of the potential anomalous time stamps in the month of April. The online NS-NMF misses all the April anomalies and the anomalies on September 13\textsuperscript{th} and January 2\textsuperscript{nd}. SNMF successfully detects the April dates. All of them misses the January 9\textsuperscript{th} anomalies which are deemed as abnormal by the operating manager and his team. The operating manager also indicate that July 7\textsuperscript{th}, July 8\textsuperscript{th}, September 17\textsuperscript{th} do not appear to be anomalies, after their extensive closer-look that yields no intelligible outcomes.  These dates are identified as anomalous by the NMF variants except NS-NMF. The operating manager registers the offline version of NS-NMF as the most competitive method followed by its online counterpart among the competitors.

\begin{table}[tb]
\centering
\caption{Most anomaly prone days identified by the four methods.}
\label{tbl:9}
\begin{tabular}{|l|}
\hline
July 4\textsuperscript{th}, 2015       \\ \hline
September 13\textsuperscript{th}, 2015 \\ \hline
September 14\textsuperscript{th}, 2015 \\ \hline
October 3\textsuperscript{rd}, 2015 \\ \hline
October 4\textsuperscript{th}, 2015 \\ \hline
October 5\textsuperscript{th}, 2015 \\ \hline
October 13\textsuperscript{th}, 2015 \\ \hline
October 14\textsuperscript{th}, 2015 \\ \hline
January 2\textsuperscript{nd}, 2016    \\ \hline
January 12\textsuperscript{th}, 2016   \\ \hline
\end{tabular}
\end{table}

Detecting the anomalies does not tell us directly the root cause behind the abnormal behaviors. But anomaly detection outcomes can be used to assign class labels to the respective data records.  A simple follow-up is to build a classification and regression tree on the labeled data sets, which can reveal which variable, or variable combination, is actually leading to these anomalous conditions. Doing so injects the interpretability to an unsupervised learning problem and can advise proper actions to address the anomalous condition and prevent future damage, disruption, or even catastrophe. An example of such exercise can be found in \cite{ahmed:2019a, ahmed:2018} and thus we will not repeat it here.

\section{Summary and Future Work} \label{sec:conclusion}
In this paper, we propose a neighborhood structure-assisted non-negative matrix factorization method and demonstrate its application in anomaly detection. We argue that in the absence of similarity information, the original NMF basis vectors are not enough to represent and separate complicated clusters in the reduced feature space. To represent and summarize the complex data structure information in a similarity matrix, we use a minimum spanning tree to capture the neighborhood connectivity information and to approximate the geodesic distance between data instances. By contrast, the alternative approaches that use Euclidean distance-based similarity is not effective, whereas the approaches using complete graphs become computationally expensive. We develop a joint optimization framework to obtain the clustering indicator and attribute distribution matrix. Then, we devise a reconstruction error-based anomaly score to flag potential point-wise anomalies. We use a parallel block stochastic gradient descent method to compute these factored matrices for fast implementation. We also design an online algorithm to render the proposed method applicable in analyzing streaming data. The specific action of modeling the neighborhood structure appears to make an appreciable impact, as NS-NMF demonstrates a clear advantage in an extensive empirical study that uses 20 benchmark data sets and one hydropower plant data set.

We believe that an important future direction is to make this NS-NMF based anomaly detection more practical in online detection. Our paper makes one of the first efforts in enabling the method online compatible. But there are still many opportunities and needs for improving the dynamic, adaptive, and robust strategies.  For instance, how to decide the optimal data buffer/block sizes, which affects the computation of anomaly scores, the frequency of score update, as well as the decision threshold.

\acks{Imtiaz Ahmed and Yu Ding were partially supported under NSF grants no. CMMI-1545038 and IIS-1849085 and under ABB project contract no. M1801386. Xia Ben Hu was partially supported by NSF grants under IIS-1750074 and IIS-1718840.}

\newpage

\appendix
\section*{Appendix}
\label{DNN}

Here, we present additional performance results with respect to two DNN methods. We choose an autoencoder with an embedding regularizer (AER)~\citep{yu:2013} and the Deep Autoencoding Gaussian Mixture Model (DAGMM)~\citep{zong:2018}. The comparison result is listed in Table~\ref{tbl:TPRDL}. NS-NMF outperforms the two DNN-based methods, producing uniquely better detection performance in 15 cases out of the 20 total. DAGMM comes out as the winner in three cases, whereas AER performs better in one case. NS-NMF and AER are tied in one case.

\begin{table*}[h]
\centering
\caption{Comparison with the DNN-based approaches}
\label{tbl:TPRDL}
\begin{tabular}{|l|l|l|l|l|}
\hline
Data set           & NS-NMF & Autoencoder & DAGMM & Winner      \\ \hline
Glass             & 4      & 0           & 0     & NS-NMF      \\ \hline
PIMA              & 8      & 4           & 1     & NS-NMF      \\ \hline
WBC               & 9      & 6           & 2     & NS-NMF      \\ \hline
Stamps            & 4      & 3           & 0     & NS-NMF      \\ \hline
Shuttle           & 2      & 0           & 0     & NS-NMF      \\ \hline
Pageblocks        & 21     & 19          & 19    & NS-NMF      \\ \hline
Heart             & 5      & 2           & 2     & NS-NMF      \\ \hline
Pendigits         & 0      & 0           & 5     & DAGMM       \\ \hline
Lymphography      & 5      & 3           & 3     & NS-NMF      \\ \hline
Waveform          & 9      & 3           & 4     & NS-NMF      \\ \hline
Cardiotocograpghy & 29     & 29          & 27    & Tie         \\ \hline
Annthyroid        & 27     & 10          & 94    & DAGMM       \\ \hline
Parkinson         & 3      & 0           & 1     & NS-NMF      \\ \hline
ALOI              & 193    & 371         & 47    & AER       \\ \hline
WDBC              & 7      & 4           & 5     & NS-NMF      \\ \hline
Ionosphere        & 92     & 63          & 79    & NS-NMF      \\ \hline
WPBC              & 11     & 10          & 9     & NS-NMF      \\ \hline
KDD               & 102    & 83          & 82    & NS-NMF      \\ \hline
Spambase          & 47     & 31          & 61    & DAGMM       \\ \hline
Arrhythmia        & 4      & 2           & 2     & NS-NMF      \\ \hline
\end{tabular}
\end{table*}

\vskip 0.2in
\bibliography{jmlr_01Jan2021}
\end{document}